\pdfoutput=1

\documentclass[11pt]{article}

\usepackage[final]{acl}
\usepackage{multirow}
\usepackage{bbding}
\usepackage{threeparttable}

\usepackage{tikz}
\usepackage{amssymb}
\usepackage{xcolor}
\usepackage{subcaption}
\usepackage{times}
\usepackage{latexsym}
\usepackage{enumitem}

\newcommand{\smallblackbox}{\raisebox{0.5ex}{\scriptsize$\blacksquare$}}

\usepackage[T1]{fontenc}

\usepackage[utf8]{inputenc}

\usepackage{microtype}
\usepackage{amsmath}
\usepackage{inconsolata}

\usepackage{graphicx}
\usepackage{booktabs}

\usepackage{framed}
\newenvironment{promptbox}{
    \definecolor{shadecolor}{rgb}{0.93,0.93,0.93}
    \begin{shaded}
}{
    \end{shaded}
}

\definecolor{darkgreen}{rgb}{0.0, 0.5, 0.0}

\title{Exploring Performance Contrasts in TableQA: Step-by-Step Reasoning Boosts Bigger Language Models, Limits Smaller Language Models}

\author{Haoyan Yang \\
  Center for Data Science \\
  \href{mailto:hy2847@nyu.edu}{\texttt{hy2847@nyu.edu}}
  \And
  Yixuan Wang \\
  Center for Data Science \\
  \href{mailto:yw7872@nyu.edu}{\texttt{yw7872@nyu.edu}}
  \And
  Keyue Tong \\
  Center for Data Science \\
  \href{mailto:kt2978@nyu.edu}{\texttt{kt2978@nyu.edu}}
  \AND
  Hongjin Zhu \\
  Center for Data Science \\
  \href{mailto:hz2291@nyu.edu}{\texttt{hz2291@nyu.edu}}
  \And
  Yuanxin Zhang \\
  Center for Data Science \\
  \href{mailto:yz6201@nyu.edu}{\texttt{yz6201@nyu.edu}}
  }

\begin{document}
\maketitle

\begin{abstract}
This paper proposes a detailed prompting flow, termed Table-Logic, to investigate the performance contrasts between bigger and smaller language models (LMs) utilizing step-by-step reasoning methods in the TableQA task. The method processes tasks by sequentially identifying critical columns and rows given question and table with its structure, determining necessary aggregations, calculations, or comparisons, and finally inferring the results to generate a precise prediction. By deploying this method, we observe a 7.8\% accuracy improvement in bigger LMs like Llama-3-70B compared to the vanilla on HybridQA, while smaller LMs like Llama-2-7B shows an 11\% performance decline. We empirically investigate the potential causes of performance contrasts by exploring the capabilities of bigger and smaller LMs from various dimensions in TableQA task. Our findings highlight the limitations of the step-by-step reasoning method in small models and provide potential insights for making improvements. The work is publicly available at \url{https://github.com/Joyyang158/NLU1012-Project}.

\end{abstract}

\section{Introduction}
Step-by-step reasoning \cite{wei2023chainofthought, dutta2024think} is a method that generates intermediate steps to facilitate logical deductions, enabling large language models (LLMs) to leverage their intermediate thoughts to generate the final answer better. The Table Question Answering (TableQA) task \cite{jin2022survey, lei2023tableqakit} involves answering questions based on tabular data, where LLMs need to perform multiple steps including locating, aggregating, and inferencing. Step-by-step reasoning has emerged as a suitable method for addressing these challenges. Previous work \cite{sui2024table, wang2024chainoftable} on TableQA using this approach mostly relies on bigger LMs as the foundation, such as GPT-4 \cite{openai2024gpt4} and PaLM-2 \cite{anil2023palm}. However, we suppose that this approach may not be suitable for smaller LMs (e.g. 7B or smaller models). In this study, we will then refer to models with parameter sizes greater than 70B as bigger LMs, and those around 7B as smaller LMs.

\begin{figure}[t]
    \centering
    \includegraphics[width=0.48\textwidth]{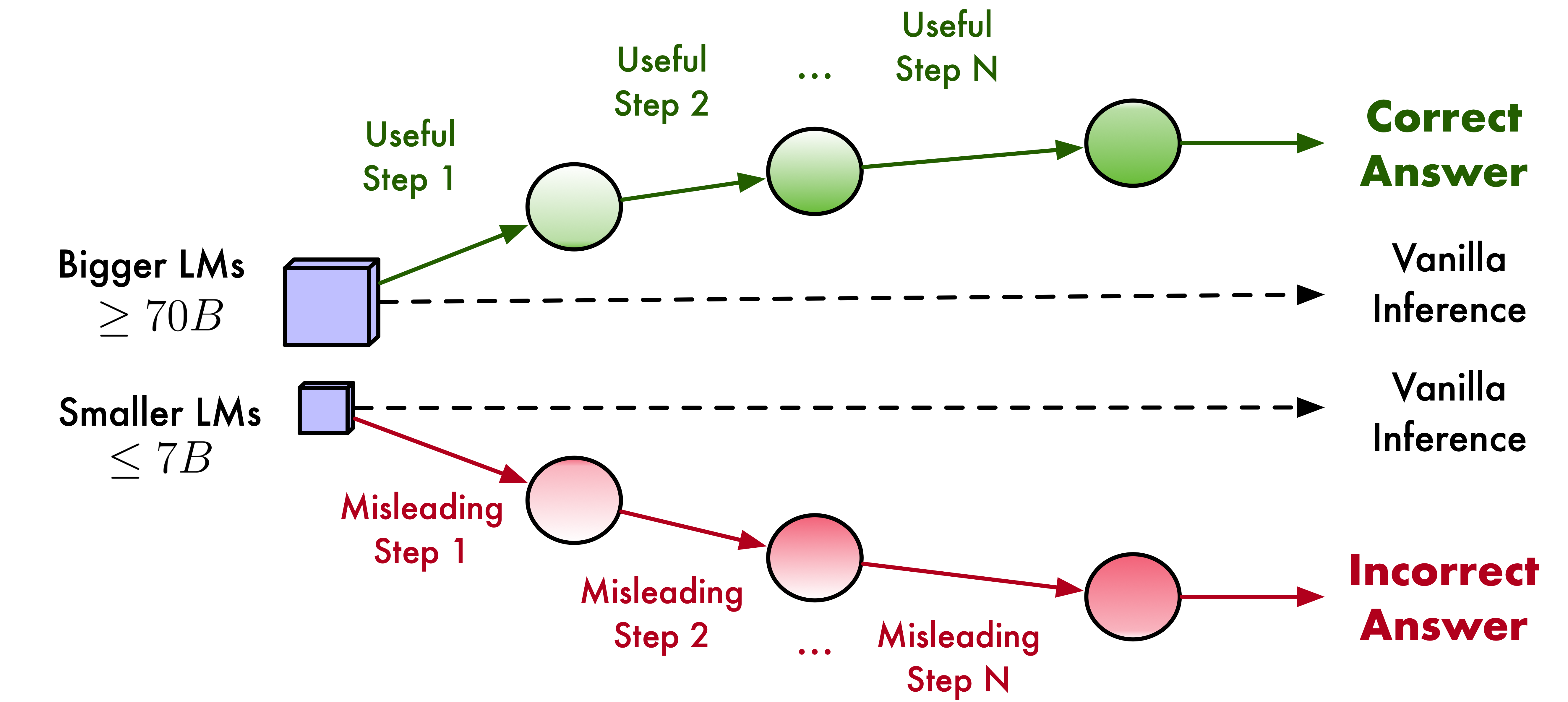}
    \caption{Comparative illustration of step-by-step reasoning paths taken by bigger and smaller LMs in a TableQA task. The diagram shows how bigger LMs typically follow a sequence of useful steps leading to accurate answers (indicated by green circles), while smaller LMs may begin with misleading steps (indicated by red circles), ultimately leading to incorrect answers.}
    \label{fig1}
    \vspace{-10pt}
\end{figure}

As shown in Figure \ref{fig1}, we analyze this inconsistency arising from the different quality of intermediate information generated by bigger and smaller LMs. For example, in the TableQA task, a possible intermediate step is to filter out the rows and columns relevant to the question. Smaller models struggle to correctly generate the answer, leading to subsequent reasoning based on misleading rows and columns, resulting in wrong results. Therefore, this paper aims to empirically explore whether the performance contrasts in step-by-step reasoning between bigger and smaller LMs for the TableQA task align with our assumptions. Additionally, we analyze the capabilities of smaller LMs needed to enhance the task to bridge the gap between bigger and smaller models.

We propose a step-by-step reasoning method named Table-logic, inspired by how humans answer questions based on tables. This method employs sequential prompting to guide the model through understanding the table, locating relevant information, performing calculations, and finally deriving the answer. Experimental results show that this approach, as expected, enhances performance in bigger LMs but deteriorates in smaller LMs. To further investigate, we designed seven sub-tasks related to the table and question, comparing the capabilities of bigger and smaller models comprehensively. This analysis provides insights into specific areas where improvements can be made to the performance of small models in this task. Overall, our main contributions of this paper are:
\begin{itemize}[label=\smallblackbox, leftmargin=0pt, itemindent=15pt]
    \item We propose a more specific and effective step-by-step reasoning method that enhances the bigger LMs' ability to handle the TableQA task.
    \item We empirically discover the performance contrasts between bigger and smaller LMs in the task, highlighting the non-generalizability of the step-by-step reasoning method for smaller models.
    \item We quantitatively and fine-grainedly compare the capabilities of bigger and smaller LMs in this task, offering valuable insights for future research to improve the performance of smaller models.
\end{itemize}

\section{Related Work}
Focusing on the TableQA task, Chain-of-thought (CoT, \citealp{wei2023chainofthought}) as a primary strategy, demonstrates that step-by-step reasoning could enhance LLMs' performance, albeit its effectiveness heavily depends on the model size. Presently, the CoT landscape offers various methods to improve TableQA tasks. For instance, \citet{zhou2023leasttomost} proposes the least-to-most prompting method, which divides the QA process into question decomposition and subproblem solving. Additionally, \citet{wang2023selfconsistency} suggests replacing the greedy decoding strategy in CoT with self-consistency, while \citet{wang2024chainoftable} introduces the chain-of-table framework to explicitly utilize tabular data, iteratively updating the table to represent tabular reasoning chains. 

While the above strategies require multiple intermediate steps, \citet{sui2024table} demonstrates that Self-augmentation, a method tailored for TableQA involving one intermediate step of pinpointing key values and ranges in a table, can boost the performance of LLMs. However, previous methods, including this one, have primarily been tested on bigger LMs, particularly on GPT 3.5 and GPT 4. \citet{huang2022large} suggests that this may be because bigger models are better at enhancing their reasoning abilities with unlabeled datasets. Motivated by the performance differences of Self-augmentation between bigger and smaller models, it is essential to investigate whether there exists more detailed segments of this ability in TableQA, examine these abilities and conduct comparative analyses between bigger and smaller LMs.

\section{Methodology}
\label{sec: methodology}
\subsection{Table-Logic Reasoning Method}

\begin{figure*}[ht!]
    \centering
    \begin{subfigure}[t]{0.21\textwidth}
        \centering
        \includegraphics[width=\textwidth]{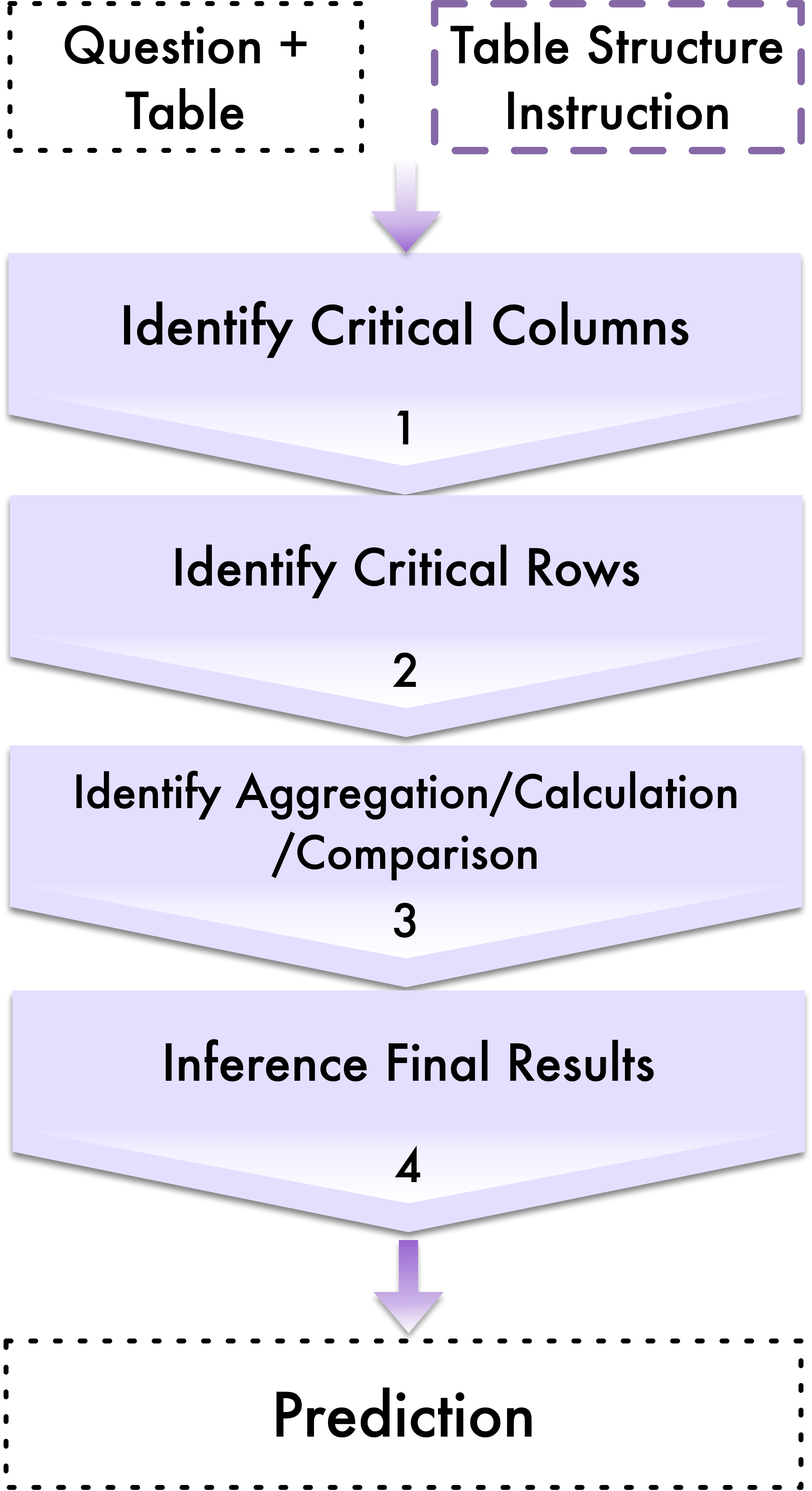}
        \captionsetup{width=1.5\textwidth}
        \caption{Process flow of Table-Logic Sequential Prompting, detailing the steps from understanding the table structure to predicting outcomes.}
        \label{fig2_sub1}
    \end{subfigure}
    \hspace{2cm}
    \begin{subfigure}[t]{0.5\textwidth}
        \centering
        \includegraphics[width=\textwidth]{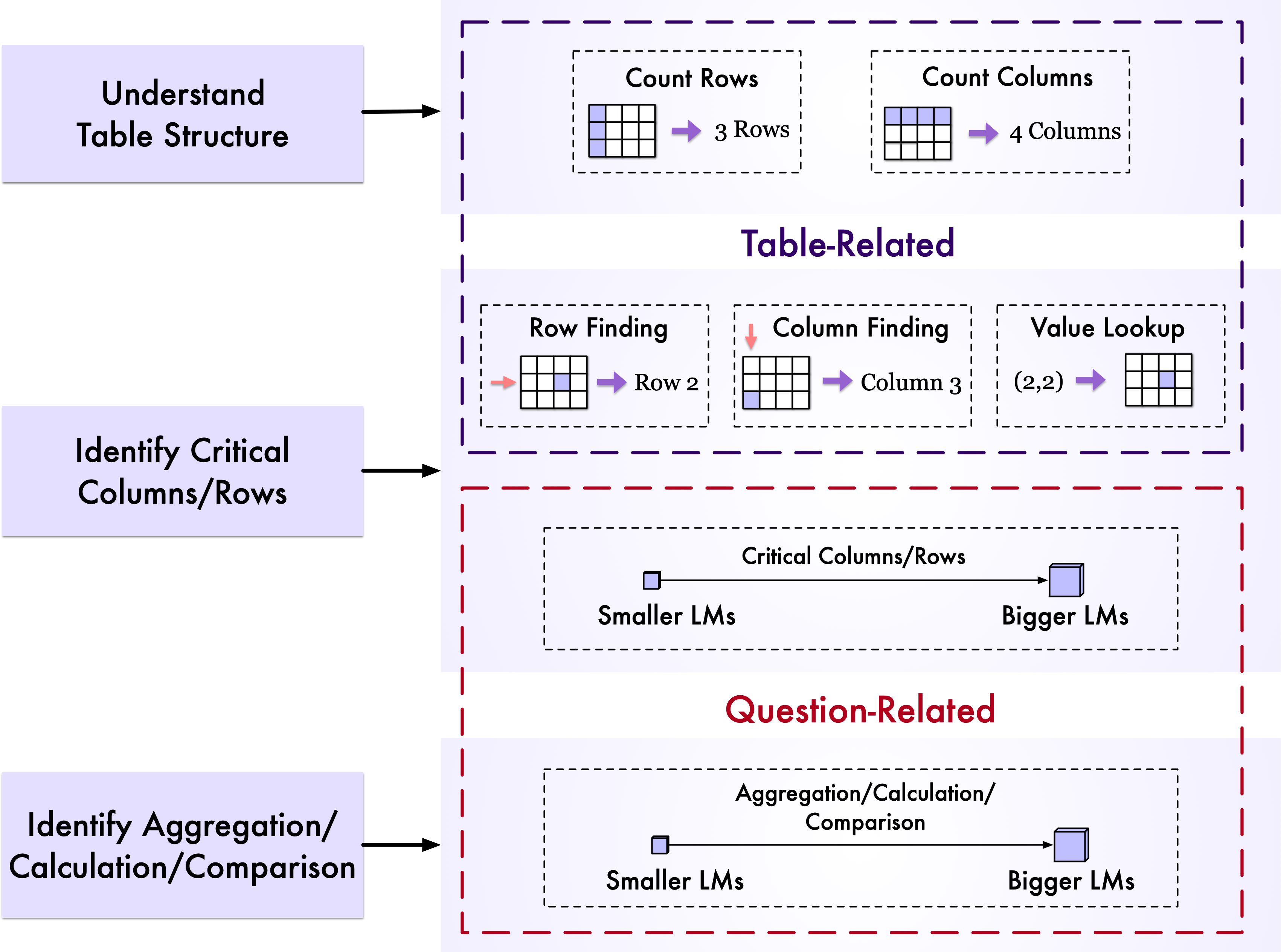}
        \caption{Sub-tasks design for analyzing the differences in handling the TableQA task between smaller and bigger LMs.}
        \label{fig2_sub2}
    \end{subfigure}
    \caption{Overview of the Table-Logic Sequential Prompting methodology and a breakdown of capabilities between smaller and bigger LMs in the TableQA task.}
    \vspace{-10pt}
    \label{fig2}
    
\end{figure*}

We propose a step-by-step reasoning method named Table-Logic, which decomposes the TableQA task into several clear steps for processing. Our approach aims to enable LLMs to generate intermediate outputs for simpler tasks through prompt, thereby avoiding the need to process and handle large amounts of information at once. As shown in Figure \ref{fig2_sub1}, the LLM sequentially identifies key columns, key rows, and any required aggregations, calculations, or comparisons to answer the question based on the table and its table structure. Note that for each step, the prompt includes a reference to the output of the previous step. The prompt of every step is shown in Appendix \ref{appendix1}.

\subsection{Sub-tasks Design}
As shown in Figure \ref{fig2_sub2}, based on the steps in the Table-Logic flow, we design seven sub-tasks to analyze the differences in the capabilities required by smaller and bigger models for handling TableQA tasks. These sub-tasks are categorized into table-related and question-related tasks, providing a comprehensive evaluation of the models' abilities based on the three aspects — understanding table structure, identifying critical columns and rows, and identifying any aggregations, calculations, or comparisons. Here are brief introductions to each sub-task:
\begin{enumerate}[leftmargin=*, label=\roman*.]
\item \textbf{Understand Table Structure}
\begin{itemize}[label=\smallblackbox, leftmargin=0pt, itemindent=15pt]
\item Count Rows: Determine the number of rows in the table.
\item Count Columns: Determine the number of columns in the table.
\end{itemize}

\item \textbf{Identify Critical Columns and Rows}
\begin{itemize}[label=\smallblackbox, leftmargin=0pt, itemindent=15pt]
\item Row Finding: Find the index of a row given a value.
\item Column Finding: Find the index of a column given a value.
\item Value Lookup: Find a value given its row and column indices.
\item Critical Columns and Rows Replacement: Apply the critical columns and rows generated by the smaller LMs to the bigger LMs' reasoning and compare the results.

\end{itemize}

\item \textbf{Identify Aggregation, Calculation or Comparison}
\begin{itemize}[label=\smallblackbox, leftmargin=0pt, itemindent=15pt]
\item Aggregation, Calculation, or Comparison Replacement: Apply the aggregations, calculations, or comparisons generated by the smaller LMs to the bigger LMs' reasoning and compare the results.
\end{itemize}
\end{enumerate}

By decomposing the TableQA task into these specific sub-tasks, we aim to evaluate and compare the performance of smaller and bigger LMs in each aspect. This detailed breakdown helps us understand where smaller LMs are lacking and what specific skills need to be improved to enhance their overall performance in TableQA tasks.

\section{Experimental Setup}
\begin{table*}[t]
\centering
\normalsize
\renewcommand{\arraystretch}{1.25}
\caption{Comparison of model performance across datasets using three different prompting methods: the experiments are conducted on two sets of models, with $>70B$ (upper section) and $\sim7B$ parameters (lower section).}
\resizebox{1\textwidth}{!}{%
\begin{tabular}{@{}c|ccc|ccc|ccc}
\hline
\multirow{2}{*} & \multicolumn{3}{c|}{Llama-3-70B} & \multicolumn{3}{c|}{GPT-3.5-Turbo} & \multicolumn{3}{c}{Qwen-1.5-72B} \\ \cline{2-10} 
                    & VAN & SA & T-Logic & VAN & SA & T-Logic & VAN & SA & T-Logic  \\ \hline
TAT-QA                  & 0.840 & 0.870\textcolor{darkgreen}{\small  ~ $\uparrow$3.0\%} & \textbf{0.874}\textcolor{darkgreen}{\small  ~ $\uparrow$3.4\%} & 0.796 & 0.816\textcolor{darkgreen}{\small  ~ $\uparrow$2.0\%} & \textbf{0.818}\textcolor{darkgreen}{\small  ~ $\uparrow$2.2\%} & 0.776 & 0.812\textcolor{darkgreen}{\small  ~ $\uparrow$3.6\%} & \textbf{0.824}\textcolor{darkgreen}{\small  ~ $\uparrow$4.8\%}  \\ 
HybridQA                  & 0.745 & 0.816\textcolor{darkgreen}{\small  ~ $\uparrow$7.1\%} & \textbf{0.823} \textcolor{darkgreen}{\small  ~ $\uparrow$7.8\%}&  0.612 & \textbf{0.673}\textcolor{darkgreen}{\small  ~ $\uparrow$6.1\%} & 0.667 \textcolor{darkgreen}{\small  ~ $\uparrow$5.5\%} & 0.648 & 0.627 \textcolor{red}{\small  ~ $\downarrow$2.1\%} & \textbf{0.654}\textcolor{darkgreen}{\small  ~ $\uparrow$0.6\%} \\
WikiTQ                 & 0.642 & 0.694\textcolor{darkgreen}{\small  ~ $\uparrow$5.2\%} & \textbf{0.710}\textcolor{darkgreen}{\small  ~ $\uparrow$6.8\%} & 0.576 & 0.608\textcolor{darkgreen}{\small  ~ $\uparrow$3.2\%}
 & \textbf{0.622}\textcolor{darkgreen}{\small  ~ $\uparrow$4.6\%} & 0.590 & 0.632\textcolor{darkgreen}{\small  ~ $\uparrow$4.2\%} & \textbf{0.648}\textcolor{darkgreen}{\small  ~ $\uparrow$5.8\%} \\ \hline
\multirow{2}{*} & \multicolumn{3}{c|}{Llama-2-7B} & \multicolumn{3}{c|}{Vicuna-7B} & \multicolumn{3}{c}{Qwen-1.5-7B} \\ \cline{2-10}
                    & VAN & SA & T-Logic & VAN & SA & T-Logic & VAN & SA & T-Logic  \\ \hline
TAT-QA                  & 0.522 & 0.442\textcolor{red}{\small  ~ $\downarrow$8.0\%} & \textbf{0.526}\textcolor{darkgreen}{\small  ~ $\uparrow$0.4\%} & \textbf{0.481} & 0.436\textcolor{red}{\small  ~ $\downarrow$4.5\%} & 0.472\textcolor{red}{\small  ~ $\downarrow$0.9\%} & \textbf{0.638} & 0.630\textcolor{red}{\small  ~ $\downarrow$0.8\%} & 0.586\textcolor{red}{\small  ~ $\downarrow$5.2\%} \\
HybridQA                &\textbf{0.443} & 0.303\textcolor{red}{\small  ~ $\downarrow$14.0\%} & 0.333\textcolor{red}{\small  ~ $\downarrow$11.0\%} & \textbf{0.300} & 0.186\textcolor{red}{\small  ~ $\downarrow$11.4\%} & 0.242\textcolor{red}{\small  ~ $\downarrow$5.8\%} & \textbf{0.603} & 0.520\textcolor{red}{\small  ~ $\downarrow$8.3\%}& 0.480\textcolor{red}{\small  ~ $\downarrow$12.3\%}\\
WikiTQ                  & \textbf{0.412} & 0.406\textcolor{red}{\small  ~ $\downarrow$0.6\%} & 0.403\textcolor{red}{\small  ~ $\downarrow$0.9\%} & \textbf{0.364} & 0.308\textcolor{red}{\small  ~ $\downarrow$5.6\%} & 0.324\textcolor{red}{\small  ~ $\downarrow$4.0\%} & 0.434 & \textbf{0.445}\textcolor{darkgreen}{\small  ~ $\uparrow$1.1\%} & 0.433\textcolor{red}{\small  ~ $\downarrow$0.1\%}\\ \hline
\end{tabular}%
}
\label{tab1}
\end{table*}

To validate our hypothesis, we select three bigger LMs ($\geq 70B$) and three smaller LMs ($\sim 7B$) and conduct experiments with Table-logic (T-Logic) alongside two other baselines across three datasets.

\subsection{Baseline Prompting Methods}
\noindent \textbf{Vanilla (VAN)}: The model is evaluated based solely on its inherent capabilities, without any additional augmentations or modifications.

\noindent \textbf{Self-Augmentation (SA)} \cite{sui2024table} :
This method involves two phases. In the first phase, self-augmented prompts are used to ask the LLM to generate additional knowledge about the table. In the second phase, the response is incorporated into a follow-up prompt to request the final answer.

\subsection{Models}
\noindent \textbf{Bigger LMs}: Llama-3-70B, GPT-3.5-Turbo, Qwen1.5-72B \cite{bai2023qwen}

\noindent \textbf{Smaller LMs}: Llama-2-7B \cite{touvron2023llama}, Vicuna-7B, QWEN1.5-7B \cite{bai2023qwen}
 
All models, except GPT, are accessed via the TogetherAI API \footnote{ \url{https://api.together.ai/models}}, while GPT-3.5-Turbo is accessed via the OpenAI API \footnote{\url{https://platform.openai.com/docs/models}}.

\subsection{Datasets}
\noindent \textbf{TAT-QA} \cite{zhu2021tat}: A QA dataset on a hybrid of tabular and textual content in finance. 

\noindent \textbf{HybridQA} \cite{chen2021hybridqa}: A QA dataset that requires reasoning over a hybrid of tabular and textual data from Wikipedia. 

\noindent \textbf{WikiTableQuestions (WikiTQ)} \cite{pasupat2015compositional}: A dataset of complex questions on Wikipedia tables. 

\subsection{Evaluation Metric}
In all experiments, we use GPT-4 to reference the gold answer and determine whether the prediction is correct, calculating accuracy as the evaluation metric. The prompt used is shown in Appendix \ref{appendix2}. Compared to traditional metrics, one advantage is its ability to understand structure, semantics, and multiple aspects of the language sentence \cite{liu2023geval, zhao2023investigating}. 

Additionally, we sample 800 examples from the three datasets to compare human evaluation results with those of GPT-4, achieving a 93.6\% agreement. This indicates that GPT-4 is a reliable evaluation tool for this task.

\subsection{Experimental Description}
We conduct comparative experiments to test whether there are performance contrasts between bigger and smaller LMs, and to evaluate whether our T-Logic method is a better step-by-step reasoning approach compared to baselines. 

Moreover, we conduct experiments on seven designed sub-tasks to quantitatively analyze the capability differences between smaller and bigger LMs from three aspects to investigate the potential reasons for performance differences. Firstly, we test the existence of incapability in smaller LMs regarding table structure, critical rows and columns, and aggregation. We incorporate useful information related to these three aspects into the vanilla models to ensure that all the capabilities we compared were areas where smaller LMs were lacking. Improving these capabilities is essential for enhancing the performance of smaller LMs on the TableQA task. The table structure is manually written, while the critical rows, columns, and aggregations are generated by the bigger model Llama-3-70B. Subsequently, We average the performance of the three bigger models and the three smaller LMs, summarizing the capability differences between bigger and smaller LMs across these seven tasks. Finally, we divide the seven sub-tasks into table-related and question-related categories and conduct a detailed analysis of the potential reasons from these perspectives.

\section{Results}

\subsection{Comparative Experiments}

\begin{table*}[t]
\centering
\small
\renewcommand{\arraystretch}{1.15}
\caption{Abilities comparison of bigger and smaller LMs across seven dimensions in the TableQA task.}
\small
\begin{tabular}{cccccccc}
\hline
& \textbf{Count} & \textbf{Count} & \textbf{Value} & \textbf{Column} & \textbf{Row} & \textbf{Critical Rows} & \textbf{Aggregation} \\
& \textbf{Rows} & \textbf{Columns} & \textbf{Lookup} & \textbf{Finding} & \textbf{Finding} & \textbf{and Columns} & \\ \hline
Bigger LMs & 0.493 & 0.957 & 0.390 & 0.717 & 0.483 & 0.737 & 0.708 \\ 
Smaller LMs & 0.173 & 0.340  & 0.083 & 0.207 & 0.157 & 0.611 & 0.606 \\
Gaps & 0.320 & 0.617 & 0.307 & 0.510  & 0.326 & 0.126 & 0.102 \\ \hline
\end{tabular}
\label{tab3}
\end{table*}

\begin{figure*}[t]
    \centering
    \small
    \begin{minipage}[t]{0.3\textwidth}
        \centering
        \includegraphics[width=\textwidth]{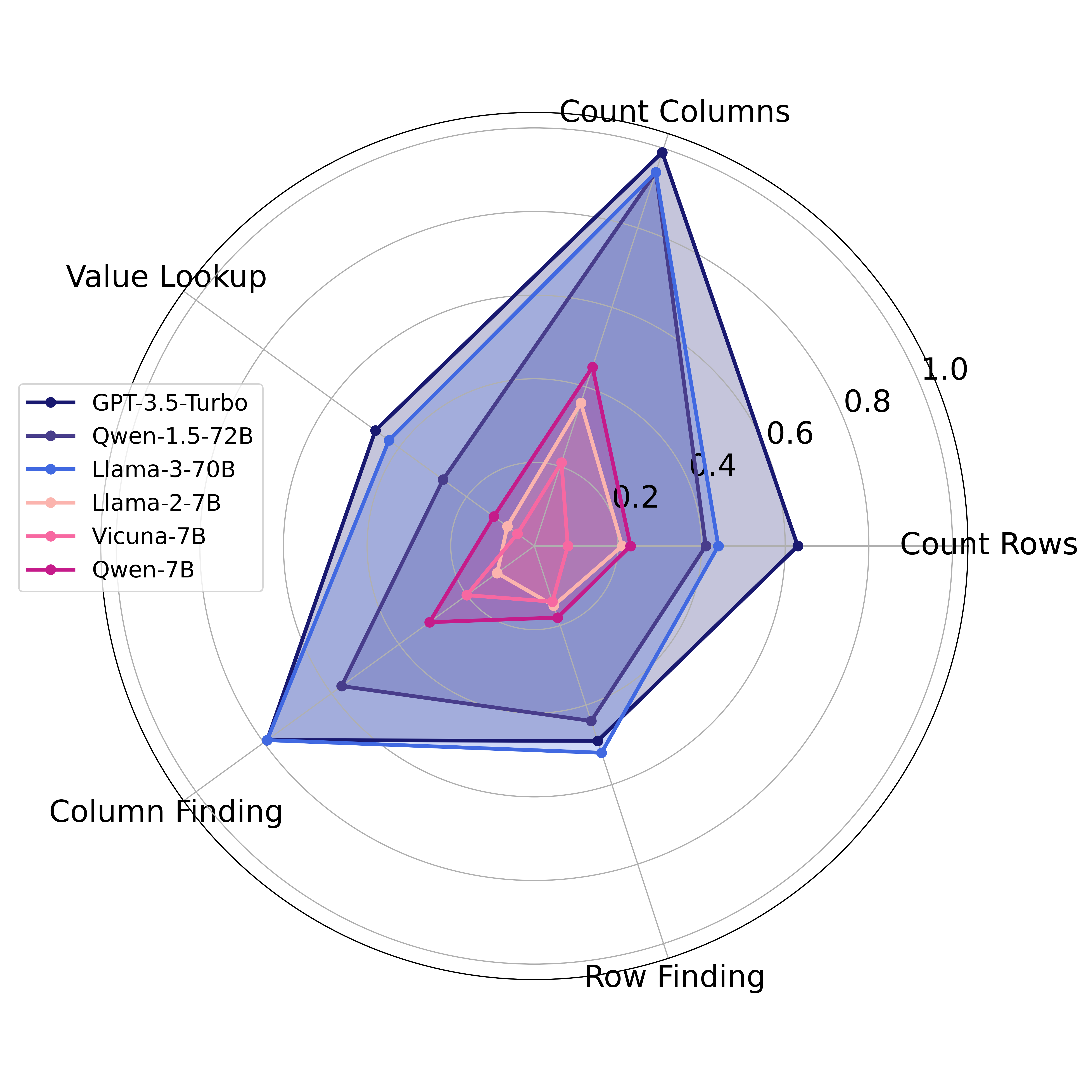}
        \caption{Radar chart displaying the table-related capabilities of bigger and smaller LMs across three datasets averagely.}
        \label{fig3}
    \end{minipage}%
    \hspace{0.1\textwidth}
    \begin{minipage}[t]{0.5\textwidth}
        \centering
        \includegraphics[width=\textwidth]{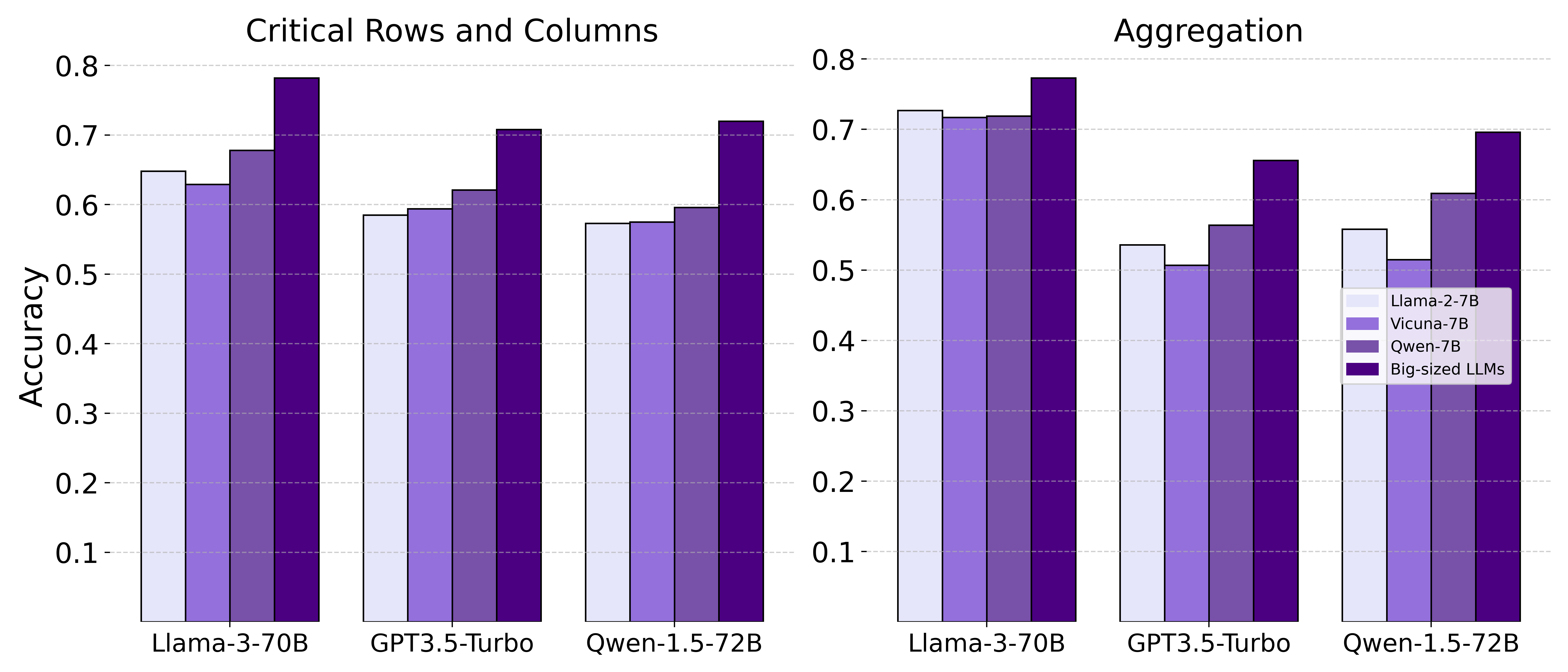}
        \caption{Bar chart displaying the question-related capabilities of smaller LMs under configurations of different bigger LMs across three datasets averagely.}
        \label{fig4}
    \end{minipage}
\end{figure*}
As demonstrated in Table \ref{tab1}, both SA and T-Logic demonstrate superior performance over vanilla methods in bigger models. However, the opposite is true for smaller LMs, where these step-by-step reasoning methods perform worse than vanilla in most cases. For example, when comparing llama-3-70B and llama-2-7B, using T-logic improves performance by 7.8\% in the bigger model but decreases it by 11\% in the smaller LMs. This result validates our hypothesis that step-by-step reasoning methods are unsuitable for smaller LMs.

In addition, T-Logic outperforms SA in most cases, indicating that the T-Logic method is more effective. We also notice that the performance drop of SA was more significant than that of T-Logic in smaller LMs. Since SA involves only two steps, we conclude that a generalized, mixed, and complex prompt is more likely to produce misleading information during intermediate steps in small models, leading to further performance degradation.

\subsection{Quantitative Analysis}

\subsubsection{Existence of Incapability}
As shown in Table \ref{tab2}, the performance of almost all smaller LMs improves after incorporating information related to the three aspects. This demonstrates that smaller LMs lack capabilities in these areas and that this phenomenon is widespread. It highlights the reasonableness and significance of studying the gap between smaller and bigger LMs from these three aspects.
\begin{table}[h]
    \centering
    \caption{Performance comparison of three smaller LMs under various configurations including instructions generated by Llama-3-70B.}
    \label{tab2}
    \resizebox{0.48\textwidth}{!}{%
    \begin{tabular}{cccccc}
    \toprule
    \textbf{Dataset} & \textbf{Model} & \textbf{Vanilla} & \textbf{With} & \textbf{With Column} & \textbf{With} \\
                   &                 &                  & \textbf{Table Structure}                      & \textbf{and Row}     & \textbf{Aggregation}                          \\
    \midrule›
    \multirow{3}{*}{TATQA}  & Llama-2-7B  & 0.522 & 0.614 & 0.563 & 0.536 \\
                          & Vicuna-7B & 0.481 & 0.574 & 0.538 & 0.494 \\
                          & Qwen-1.5-7B   & 0.638 & 0.675 & 0.680 & 0.649 \\
    \midrule
    \multirow{3}{*}{HybridQA} & Llama-2-7B  & 0.443 & 0.477 & 0.541 & 0.461 \\
                            & Vicuna-7B & 0.300 & 0.294 & 0.423 & 0.310 \\
                            & Qwen-1.5-7B  & 0.603 & 0.586 & 0.619 & 0.592 \\
   
    \midrule
    \multirow{3}{*}{WikiTQ} & Llama-2-7B  & 0.412 & 0.445 & 0.665 & 0.422 \\
                          & Vicuna-7B & 0.364 & 0.435 & 0.626 & 0.379 \\
                          & Qwen-1.5-7B   & 0.434 & 0.420 & 0.670 & 0.455 \\
    \bottomrule
    \end{tabular}}
\end{table}

\subsubsection{Gaps between bigger and smaller LMs}
Table \ref{tab3} shows the average performance of bigger and smaller LMs on seven tasks, with the largest performance gaps observed in the count columns and column finding tasks. We analyze these gaps by categorizing the tasks into table-related and question-related tasks. 

As shown in Figure \ref{fig3}, in table-related tasks, all three smaller models perform significantly worse than the bigger models, with the largest gap in the column finding task, reaching 78\%. Each smaller model had specific weaknesses; for example, Vicuna-7B struggles with counting columns, while Llama-2 performs poorly in column finding.

As shown in Figure \ref{fig4}, for the question-related tasks, the capability gap between smaller and bigger LMs was not as pronounced as in table-related tasks. This suggests that smaller models are more likely to treat tables as paragraphs of information rather than understanding their structure. As a result, while smaller models can identify key information based on the questions (such as critical contents), they struggle to understand how the information corresponds to a specific location in the table.

These analyses investigate the potential reasons why step-by-step reasoning is ineffective for smaller models in the TableQA task. Therefore, to improve the performance of smaller models in the task, it is essential to enhance their ability to understand table structures firstly, such as recognizing the number of columns and identifying the corresponding headers.

\section{Conclusion}
In conclusion, our research empirically demonstrates the performance contrasts between bigger and smaller LMs on the TableQA task when using a step-by-step reasoning approach based on our proposed Table-Logic prompting method. Through comprehensive analysis across various dimensions, we denote the reasons behind its ineffectiveness on smaller LMs. Our study provides deeper insights into the limitations of applying advanced prompting methods in the TableQA task to smaller LMs.

\section*{Limitation}
\begin{itemize}[label=\smallblackbox, leftmargin=0pt, itemindent=15pt]
\item Table-logic reasoning method requires multiple calls to LLMs, which can be costly. 
\item  We do not consider the interaction effects between model capabilities in our experiments.
\end{itemize}

\bibliography{custom}

\newpage
\onecolumn
\appendix
\section{Table-Logic Flow Prompt}
\label{appendix1}
\begin{promptbox}
\label{prompt}
\noindent \textbf{1. Table reading instruction:} Table data is structured in a dictionary with keys 'header', 'rows', and 'name'. The 'header' contains column names, 'rows' includes sublists for each row matching the header, and 'name' provides the table's identifier.

\noindent\textbf{2. Identify critical columns: } Table: \{table\} Reading instruction: \{format\} Based on the reading instruction of this table, identify critical columns highly relevant to this question: \{question\}. 

\noindent\textbf{3. Identify critical rows: } Table: \{table\} Reading instruction: \{format\} Critical columns: \{column\} Based on the reading instruction and critical columns of this table, identify critical rows highly relevant to this question: \{question\}.

\noindent\textbf{4. Identify aggregation/calculation/comparison: } Table: \{table\} Reading instruction: \{format\} Critical columns: {column} Critical rows: \{row\} Based on the reading instruction and key values in critical columns and rows of this table, identify any aggregation, calculation, and comparison required by this question: \{question\}

\noindent\textbf{5. Achieve final prediction: } Table: \{table\} Reading instruction: \{format\} Given this table and read instruction, answer this question: \{question\}. You do not need to explain the answer. Additional information that may help is given below. Critical columns: \{column\} Critical rows: \{row\} Aggregation, calculation, and comparison: \{aggregation\}

\end{promptbox}

\section{Template of GPT-4 Evaluation}
\label{appendix2}

\textbf{Prompt}: You are now an intelligent assessment assistant. Based on the question and the golden answer, judge whether the predicted answer correctly answers the question and give only a Yes or No.\\
\textbf{Question}:\\
\textbf{Gold Answer}:\\
\textbf{Predicted Answer}:\\
\noindent\rule{\textwidth}{0.4pt}
\textbf{Expected Output}: Yes / No

\end{document}